\documentclass[sn-mathphys]{sn-jnl}

\usepackage{comment}
\usepackage{color}
\usepackage{xcolor}
\usepackage{amsmath}
\usepackage{amsfonts}

\newcommand{\omittext}[1]{}

\usepackage{xcolor, soul}
\sethlcolor{yellow}

\jyear{2023}%

\theoremstyle{thmstyleone}%
\theoremstyle{thmstyletwo}%

\theoremstyle{thmstylethree}%

\begin{document}

\title[Force Sensing Surgical Drill]{A force-sensing surgical drill for real-time force feedback in robotic mastoidectomy}


\author[1]{\fnm{Yuxin} \sur{Chen}}
\equalcont{These authors contributed equally to this work.}

\author[1]{\fnm{Anna} \sur{Goodridge}}
\equalcont{These authors contributed equally to this work.}

\author*[1]{\fnm{Manish} \sur{Sahu}}\email{manish.sahu@jhu.edu}
\equalcont{These authors contributed equally to this work.}

\author[1]{\fnm{Aditi} \sur{Kishore}}

\author[1]{\fnm{Seena} \sur{Vafaee}}

\author[1]{\fnm{Harsha} \sur{Mohan}}

\author[1]{\fnm{Katherina} \sur{Sapozhnikov}}

\author[1,2]{\fnm{Francis} \sur{Creighton}}

\author[1,2]{\fnm{Russell} \sur{Taylor}} 

\author[1,2]{\fnm{Deepa} \sur{Galaiya}}

\affil[1]{\orgdiv{Laboratory for Computational Sensing and Robotics}, \orgname{Johns Hopkins University}, \orgaddress{\city{Baltimore}, \state{MD}, \country{U.S}}}

\affil[2]{\orgdiv{Department of Otolaryngology–Head and Neck Surgery}, \orgname{Johns Hopkins University School of Medicine}, \orgaddress{\city{Baltimore}, \state{MD}, \country{U.S}}}


\abstract{
\textbf{\\* Purpose}:
Robotic assistance in otologic surgery can reduce the task load of operating surgeons during the removal of bone around the critical structures in the lateral skull base.
However, safe deployment into the anatomical passageways necessitates the development of advanced sensing capabilities to actively limit the interaction forces between the surgical tools and critical anatomy.
\textbf{\\* Methods}: 
We introduce a surgical drill equipped with a force sensor that is capable of measuring accurate tool-tissue interaction forces to enable force control and feedback to surgeons.
The design, calibration and validation of the force-sensing surgical drill mounted on a cooperatively controlled surgical robot are described in this work.
\textbf{\\* Results}: 
The force measurements on the tip of the surgical drill are validated with raw-egg drilling experiments, where a force sensor mounted below the egg serves as ground truth.
The average root mean square error (RMSE) for points and path drilling experiments are 41.7 ($\pm$ 12.2) mN and 48.3 ($\pm$ 13.7) mN respectively.   
\textbf{\\* Conclusions}: 
The force-sensing prototype measures forces with sub-millinewton resolution and the results demonstrate that the calibrated force-sensing drill generates accurate force measurements with minimal error compared to the measured drill forces.
The development of such sensing capabilities is crucial for the safe use of robotic systems in a clinical context.
}

\keywords{Force sensing, Surgical instruments, Mastoidectomy, Surgical robotics, Cooperative control}



\maketitle

\section{Introduction}\label{Introduction}

\begin{figure}[b]%
\centering
\includegraphics[width=0.9\textwidth]{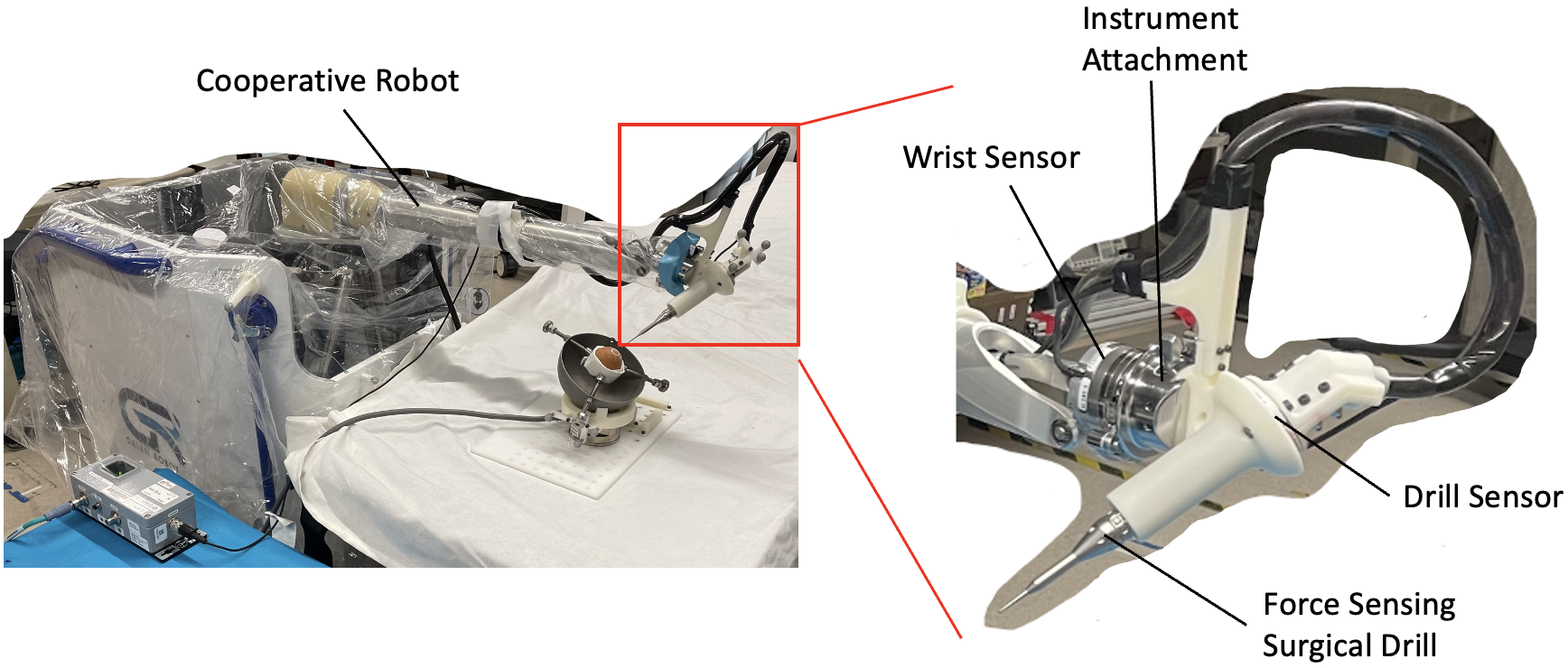}
\caption{Force sensing system mounted on a cooperatively controlled robot and containing a conventional surgical drill. }\label{overall_setup_labeled}
\end{figure}

Microsurgical interventions often involve performing fine surgical tasks in a small and delicate environment.
Microsurgery introduces additional challenges for operating surgeons since it requires the manipulation of surgical instruments near critical structures while coping with the effect of human sensory-motor limitations.

Robotic assistance can provide precise, tremor-free manipulation of surgical instruments.
However, safe deployment into the anatomical passageways requires the robot to use its sensing capabilities to control or limit tool-tissue forces  \cite{simaan2015intelligent}. 
Current available (teleoperated) robotic systems generally do not offer accurate tactile force feedback, so that surgeons rely purely on visual feedback and prior experience to limit the forces exerted onto organs and tissues.
Moreover, the forces between the microsurgical instrument and the patient's anatomy often lie below the surgeon’s sensory threshold \cite{trejos2010force}.
Thus, the safe use of these robotic systems hinges on implementing advanced sensory and control capabilities to overcome these challenges.\cite{taylor2016medical,taylor1999steady}

The aim of this work is to develop smart instruments that augment human perception by measuring tool-tissue forces to enable force feedback during robotically-assisted surgery.
Motivated by the flexibility and benefits of cooperative control, a specialized force sensing drill holder is designed for a cooperatively controlled steady-hand surgical robot.
A major issue for the development of accurate tool-tissue force measurement is to compute the exact forces exerted at the tip of the instrument.
During a drilling task, the force sensor measures not only the tool-tissue force, but also external forces such as the gravitational force and hand forces (for a robot-mounted drill).
Computational elimination of non-contact forces from F/T sensor measurements is necessary to obtain the tool-tissue force. \cite{kubus2008improving}
This study describes the development and calibration pipeline of a force-sensing drill (1) for its self-weight (gravitational) force and (2) for the surgeon’s hand force acting on the drill, both of which are required for accurate measurement of the forces at the tip of the surgical drill and are necessitated by the use of a cooperatively controlled robot.
The force sensing drill was extensively validated with raw eggshell (fine) and temporal bone phantom (coarse) drilling experiments, where an additional force sensor holding the raw egg or phantom served as ground truth.
The forces acting during the egg drilling experiment is analogous to drilling the deep-seated structures of the skull base procedures, while the course drilling experiment are analogous to initial stages of mastoidectomy when the cortical bone is removed. Our experimental results demonstrate that the calibrated force sensing drill generates accurate force measurements with minimal error compared to the measured drill forces.

In order to demonstrate the utility of a cooperative robot equipped with a force sensing drill, we compared the force measurements of our proposed setup with freehand drilling.
The similarity in force values during eggshell drilling experiments show that the proposed system does not greatly alter the drilling task, indicating its potential to be employed in a clinical context.

\section{Related Work}\label{Related Work}

Multiple works (e.g.,\cite{trejos2010force}) have introduced force sensing surgical instruments. For the sake of brevity, we will focus on force-sensing designs for surgical robots. 
Sang et al.~\cite{sang2017new} developed a robotically assisted surgical drill held by a da Vinci robot end effector with force sensing for otologic surgery. To measure the force applied on the drill tip, they calibrated for the gravitational force of the instrument, the inertial force and torque from the drill dynamics, and linear force and torque.
Su et al.~\cite{su2020deep} integrated a force sensor between the robot end effector and the surgical instrument, and developed a tool dynamics identification method to achieve accurate force sensing for bilateral teleoperation.

Within the context of ENT surgery, Rothbaum et al.~\cite{rothbaum2002robot, rothbaum2003task} evaluated the potential of robotic assistance in stapedotomy and showed that robot assistance can significantly limit the maximum force applied to the stapes footplate. 
Some of their evaluation methods inspired our experiment validation procedure. To measure and validate the force measurement on the instrument tip, we followed a similar design as described in~\cite{miroir2010robotol} and used the second force sensor measurement as ground truth.

RobOtol\cite{miroir2010robotol}, MMS\cite{maier2010mms}, and the smart tool point~\cite{taylor2010sensory} are specifically designed for ENT applications, but use the teleoperation paradigm. MMS uses a control console with two joysticks while RobOtol uses a Phantom Omni joystick to enable force feedback, and the smart tool point is controlled via handheld remote. However, teleoperation suffers from several challenges such as removing the surgeon from direct patient access and requiring significant changes to conventional surgical procedures. Cooperative systems on the other hand are cost-effective, offer intuitive control over teleoperation, and require minimal change to surgical procedures. They offer the added benefit of keeping surgeons at the bedside while requiring less training to become familiar with the robotic system.

This work reports development, calibration, and initial experiments with a force-sensing surgical drill suitable for use with a hand-over-hand cooperatively controlled surgical robot. The design gives accurate, real-time tool-to-tissue measurements while the system is attached to the robot and the surgeon is using hand-over-hand guidance of the drill. 
However, this design requires that a means be provided to distinguish between tool-tissue and hand-tool forces. 
Our design provides substantial decoupling of hand-tool forces from tool-tissue forces by using two force/torque sensors and provides efficient calibration methods for the influence of tool weight on the sensors, as well as for any residual interaction between surgeon hand forces and tool-tissue force measurements.

\section{Methods}\label{Methods}

\subsection{Cooperative Robot}\label{Sec:Galen}
The Robotic ENT Microsurgery System (REMS) is a cooperatively-controlled robot consisting of a gantry arm with five actuated degrees of freedom developed by Olds, {\it et al.} \cite{olds2014preliminary,olds2015phd}.  For this study, we are using a pre-clinical version developed by Galen Robotics (Galen Robotics, Baltimore, MD).
The robot uses an admittance control system and its kinematic design allows the precise location of the attached surgical instrument to be known with a high degree of fidelity. Using a 6-axis force/torque (F/T) sensor at the distal end of the arm, referred to here as the ``wrist" sensor, the force exerted on the handle of the instrument by the surgeon is measured and used to control the robot motion \cite{olds2014preliminary}.

\subsection{Mechanical Design}\label{Mechanical_Design}
\begin{figure}[t]%
\centering
\includegraphics[width=.90\textwidth]{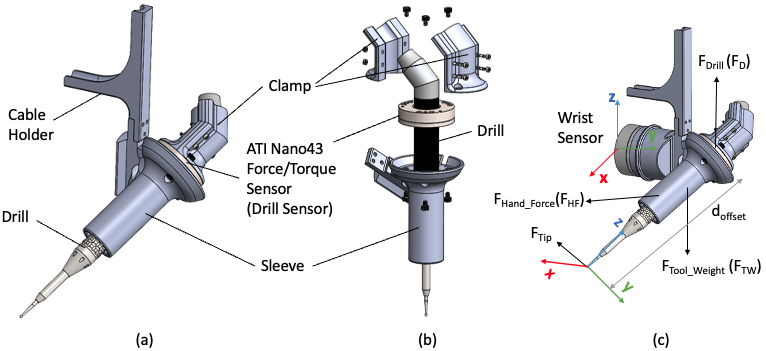}
\caption{Force Sensing Surgical Drill Design; (a) Main View; (b) Exploded View; (c) Force Applied on the Drill}\label{FSD_design}
\end{figure}

The main design objectives of our force-sensing robotic drill attachment were to decouple the hand forces on the drill used by the surgeon to control robot motion from the drill-to-tissue forces.
The system (shown in Fig.~\ref{overall_setup_labeled} and Fig.~\ref{FSD_design}) comprises an Anspach EG1 surgical drill and a 6-axis ATI Nano43 F/T sensor. Because the wrist sensor on the robot cannot distinguish between the hand force applied to the instrument and tissue force acting on the tip of the instrument, a second F/T sensor (referred to here as the ``drill" sensor) was incorporated into the design.
Based on the work by Dillon {\it et al.}~\cite{dillon2013experimental},  we expected tool-tip forces no larger than 5 N for drilling during a mastoidectomy and  much lower forces for the delicate drilling of critical structures near the facial nerve, brain and dura, blood vessels, and stapes footplate.
The drill sensor was chosen for its through-hole that allows for axial assembly onto the drill and for its sensing range and resolution, up to 18 N of force and 250 Nmm of torque with resolutions of 0.004 N and 0.05 Nmm respectively. For all experiments, the drill sensor and phantom holder sensor were sampling at 100 Hz and the wrist sensor was sampling at 200 Hz.

The assembly comprises three main components. A two-part clamp is rigidly attached to the drill at the distal end of the sensor. The sleeve that the surgeon holds is fixed to the proximal end of the sensor and fits around the drill such that no hand force will be imparted on the drill itself. The sleeve also has a dovetail that mates with the robot’s end-effector so that the hand force applied to the sleeve is measured by the wrist sensor and used to control the robot. The angle of the dovetail was designed to hold the drill in a ``neutral” position for a right-handed surgeon when the robot is at its home position, based on surgeons’ ergonomic feedback. The third component is the cable holder, mounted to the sleeve, which allows for strain relief of the drill and sensor cables as well as consistency in the position of the cables through robot travel. 

\begin{figure}[t]%
\centering
\includegraphics[width=0.9\textwidth]{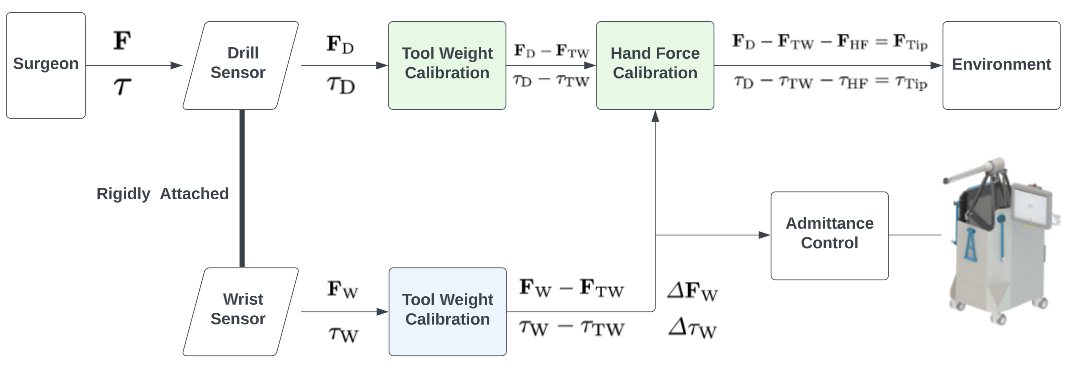}
\caption{Block diagram of whole system including force and torque from the weight of the drill, hand force of the surgeon holding the drill, and the external force acting on the drill tip. All F/T measurements are in the drill sensor coordinate frame.}\label{Diagram_of_whole_system}
\end{figure}

\subsection{Calibration}\label{Calibration}

Our calibration accounts for gravitational forces and torques on the drill force sensor arising from the tool's weight, together with residual forces coming from imperfectly shielded hand forces.  In the procedures described below, all F/T measurements are in the drill sensor coordinate frame.

\subsubsection{Tool Weight  Calibration}\label{Self-weight_Calibration}

The first step in the calibration pipeline is accounting for the gravitational force of the instrument on the drill sensor and the wrist sensor at all orientations. 

The first method used for tool weight calibration was a physics-model based approach using the known mass of the assembly, the approximated center of mass, and the known joint angles of the robot.

\begin{equation}
\mathbf{F}_{\mathrm{TW}} = R(roll, tilt) \cdot mg
\end{equation}
\begin{align}
\boldsymbol\tau_{\mathrm{TW}} &= skew(C) R(roll, tilt) mg \\
&= \begin{bmatrix}
0 & -Cz & Cy\\
Cz & 0 & -Cx \\
-Cy & Cx & 0
\end{bmatrix} R(roll, tilt) mg
\end{align}

\noindent where m is the mass of the drill; g is the acceleration due to gravity; and R(roll, tilt) is the rotation matrix of the drill sensor coordinate frame in the world coordinate frame. C = [Cx, Cy, Cz] is the vector from the origin of the drill sensor coordinate frame to the centroid of the drill. 

For the second (model-free) approach, the calibration process is realized by acquiring pairwise data of the sensor’s raw signal output for different orientations of the surgical drill and establishing the mapping by using polynomial least squares fitting in the Bernstein basis. Given pairwise data $\{ x_i, f_i \}_{i=1}^{N}$ where, $x_i = (roll_i, tilt_i)$ and $f_i$ denote the $i$-th pair of robot orientations and corresponding sensor readings respectively, and $P(x) = c_0 + c_1 x + \cdots + c_l x^n$ representing Berstein polynomial of degree $n$: The coefficients $c_j$ of the polynomial P(x) are obtained by minimizing the sum of the squares of the deviations on data.

\begin{align}
P(x_i) &=\sum_{j=0}^{n} \sum_{k=0}^{n} c_{jk} A_{j, n}\left(roll_i\right) B_{k, n}\left(tilt_i\right) \\
&= \left[A_{0, n}(roll_i), \cdots, A_{n, n}(roll_i)\right]\left[\begin{array}{ccc}
c_{00} & \cdots & c_{0 n} \\
\vdots & \ddots & \vdots \\
c_{n 0} & \cdots & c_{n n}
\end{array}\right]\left[\begin{array}{c}
B_{0, n}(tilt_i) \\
\vdots \\
B_{n, n}(tilt_i)
\end{array}\right] \label{Berstein_polynomial}
\end{align}

Where $A_{j, n}\left(roll_i\right)$ and $ B_{k, n}\left(tilt_i\right)$ are n$^{th}$ order Bernstein polynomials.  

\begin{equation}
    A_{j, n}(roll_i)=\left(\begin{array}{c} n \\ j \end{array}\right) roll_i^j (1-roll_i)^{n-j}, \quad j=0, \ldots, n
\end{equation}
\begin{equation}
    B_{k, n}(tilt_i)=\left(\begin{array}{c} n \\ j \end{array}\right) tilt_i^k (1-tilt_i)^{n-k}, \quad k=0, \ldots, n
\end{equation}

\subsubsection{Hand Force Calibration}\label{Hand_Force_Calibration}
Unlike teleoperated robots, cooperative control necessitates external force on the instrument to drive the robot. These forces are measured by the ``wrist" sensor at the end effector of the robot.  In principle, the design of our force sensing drill should shield the drill force sensor from these hand forces.  However, we observed some interactions, which are likely due to elasticity of our rapid-prototyped mechanical components and possible small compliant motions of the robot structure.  Accordingly, we implemented a second calibration step to mitigate these effects.
Using pairwise data from both the drill and wrist sensors, we built a regression model to estimate the impact of hand-to-drill forces on the drill-to-tissue force measurements as seen in the following equation:
\begin{align}
\left[\begin{array}{c}
\mathbf{F}_{\mathrm{HF}} \\
\boldsymbol\tau_{\mathrm{HF}}
\end{array}\right]= M \left[\begin{array}{c}
\Delta \mathbf{F}_{\mathrm{W}} \\
\Delta \boldsymbol\tau_{\mathrm{W}}
\end{array}\right]
\end{align}

\noindent where 
$ \Delta \mathbf{F}_{\mathrm{W}} / \Delta \boldsymbol\tau_{\mathrm{W}} $ is the change of the force and torque measurement from the wrist sensor and M represents the model of the relationship between the force and torque measured in the wrist sensor coordinate frame and in the drill sensor coordinate frame. 

Given pairwise sensor measurements data $\{ f_i^{d}, f_i^{w} \}_{i=1}^{N}$ where, $f_i^{d}$ and $f_i^{w}$ denote the $i$-th pair of drill and wrist sensor readings respectively, the regression model $M(f^w)$ is obtained by optimizing the squared error.

\begin{equation}
\sum_{i=1}^{N}  \mid M (f_i^w) - f_i^d \mid^2
\end{equation}

In this work, we evaluate the suitability of linear and non-linear (MLP, RF) regression methods for hand-force calibration.

\subsubsection{Tip Force Computation}\label{Resultant Force at Drill Tip}

Finally, the resultant force acting on the tip is computed using the forces and torques at the drill sensor.

\begin{equation}
\mathbf{f}_{tip} = [\boldsymbol\tau_{\mathrm{x}} / \mathbf{d}_{\mathrm{offset}}, \boldsymbol\tau_{\mathrm{y}} / \mathbf{d}_{\mathrm{offset}}, \mathbf{f}_{\mathrm{z}}]^{T} \\
\end{equation}

\noindent where $\boldsymbol\tau_{\mathrm{x}}, \boldsymbol\tau_{\mathrm{y}}$ are the torques about the x and y axes in drill sensor coordinates, $ \mathbf{d}_{\mathrm{offset}} $ is the distance between the drill sensor coordinate frame and the drill tip, and  $\mathbf{f}_{\mathrm{z}} $ is the force along z axis in drill sensor coordinates.


\section{Experimental Results and Discussion}\label{Experiments and results}

\subsection{Tool Weight Calibration Validation Experiments}\label{Self-weight_Calibration_Validation_Experiments}

\begin{figure}[h]%
\centering
\includegraphics[width=0.8\textwidth]{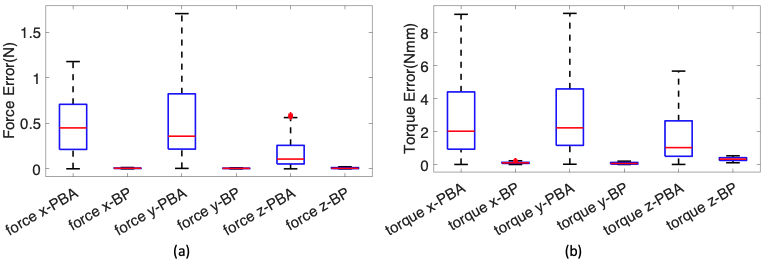}
\caption{Result of Tool Weight Calibration on the Drill Sensor Measurement Using Physics-Based Approach(PBA) and Bernstein Polynomial(BP); (a) Force Error; (b) Torque Error}\label{SW_calibration_result}
\end{figure}

For the tool weight calibration, data was collected at different orientations of the robot, covering the entire range of roll and pitch angles.
The performance comparison of physics-based and BP-based least square fit methods is shown in Fig.~\ref{SW_calibration_result}.
Most of the error seen in the physics-based approach can be attributed to modeling the instrument as a rigid body when in fact there was variation in the center of mass due to the movement in the slack of the drill and sensor cables. These variations are highest when the cables started to bend which typically happened during tilt stage movement. Cable slack was required for full robot travel, however, the changes in cable position and therefore the effect of their mass on the sensors were a large source of error. 
The BP-based least squares fit showed much less error, on the order of 0.01N and 0.2 Nmm for force and torque respectively.

\subsection{Hand Force Calibration Validation Experiments} \label{Hand_Force_Calibration_Validation_Experiments}
The data for hand force estimation was collected at different robot orientations. The robot's actuators were disabled, the drill was gently held, and force was applied in different directions to mimic the hand force that surgeons would apply to the instrument during surgery. The procedure explained in section 3.3.2 is then completed. A grid search was performed on MLP and RF to find the best parameters to fit the training data and then tested on the validation data.
The performance of different regression methods on a disjunct dataset is shown in Table~\ref{Hand Force Calibration Validation Result}.
The results suggest that non-linear regression models work slightly better than linear regression approaches. Of the two non-linear approaches, MLP works better than RF, especially in torque values, so MLP regression was the method selected for further experiments and analysis. Force in the Z direction and torque in the X and Y directions are later used for calculating the tip force and are therefore the most relevant.

\begin{table}[h]
\begin{center}
\begin{minipage}{\textwidth}
\caption{Validation Result of Hand Force Calibration}\label{Hand Force Calibration Validation Result}
\begin{tabular*}{\textwidth}{@{\extracolsep{\fill}}lcccccc@{\extracolsep{\fill}}}
\toprule%
& \multicolumn{6}{@{}c@{}}{Percent Improvement in RMSE Before and After Hand Force Calibration} \\\cmidrule{2-7}%
				 & $\mathbf{f}_{\mathrm{x}}(N)$ & $\mathbf{f}_{\mathrm{y}}(N)$ & $\mathbf{f}_{\mathrm{z}}(N)$ & $\boldsymbol\tau_{\mathrm{x}}(Nmm)$ & $\boldsymbol\tau_{\mathrm{y}}(Nmm)$ & $\boldsymbol\tau_{\mathrm{z}}(Nmm)$ \\
\midrule
Linear & 50.23\% & \textbf{67.89}\% & 2.44\% & 84.53\% & 62.12\%	& 75.53\%\\
RF & 49.13\% & 65.07\% & \textbf{3.79}\% & 83.06\% & 73.48\% & 74.76\%\\
MLP & \textbf{53.80}\% & 67.00\%	 & 3.04\% & \textbf{86.22}\% & \textbf{74.04}\% & \textbf{75.83}\%\\
\botrule
\end{tabular*}
\footnotetext{Note: Each value was calculated by taking the average of three trials; A comparison of Linear Regressor, Multi-layer Perceptron Regressor (MLP), and Random Forest Regressor (RF).}
\end{minipage}
\end{center}
\end{table}

\subsection{Drilling Validation Experiments} \label{Drilling_Validation_Experiments}

\begin{figure}[b]%
\centering
\includegraphics[width=1.0\textwidth]{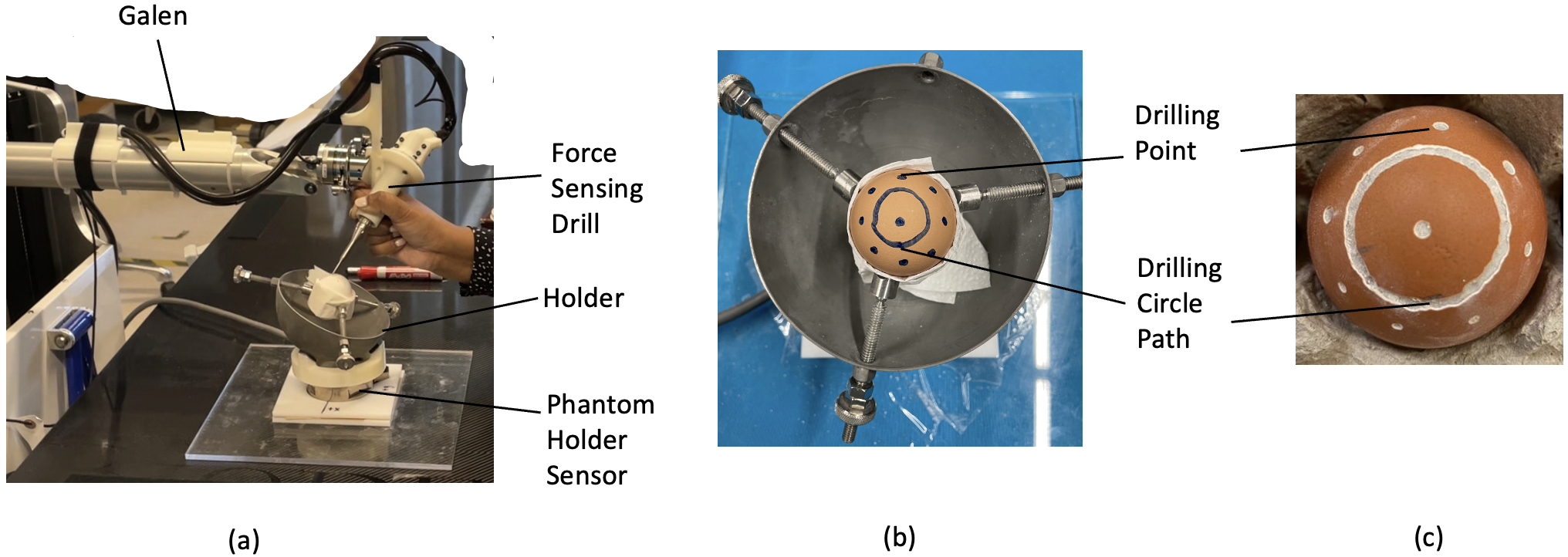}
\caption{(a) Drilling experiment setup showing surgeon holding the drill attached to the robot. The phantom holder sensor is a 6-axis ATI Gamma F/T sensor used as ground truth for tool tip force. (b) and (c) show the egg before and after the drilling experiment, respectively.}\label{Egg_drilling_experiments_setup}
\end{figure}

\subsubsection{Egg Drilling Experiment} \label{Egg_Drilling_Experiment_Procedure_and_Result}
\begin{figure}[h]%
\centering
\includegraphics[width=0.90\textwidth]{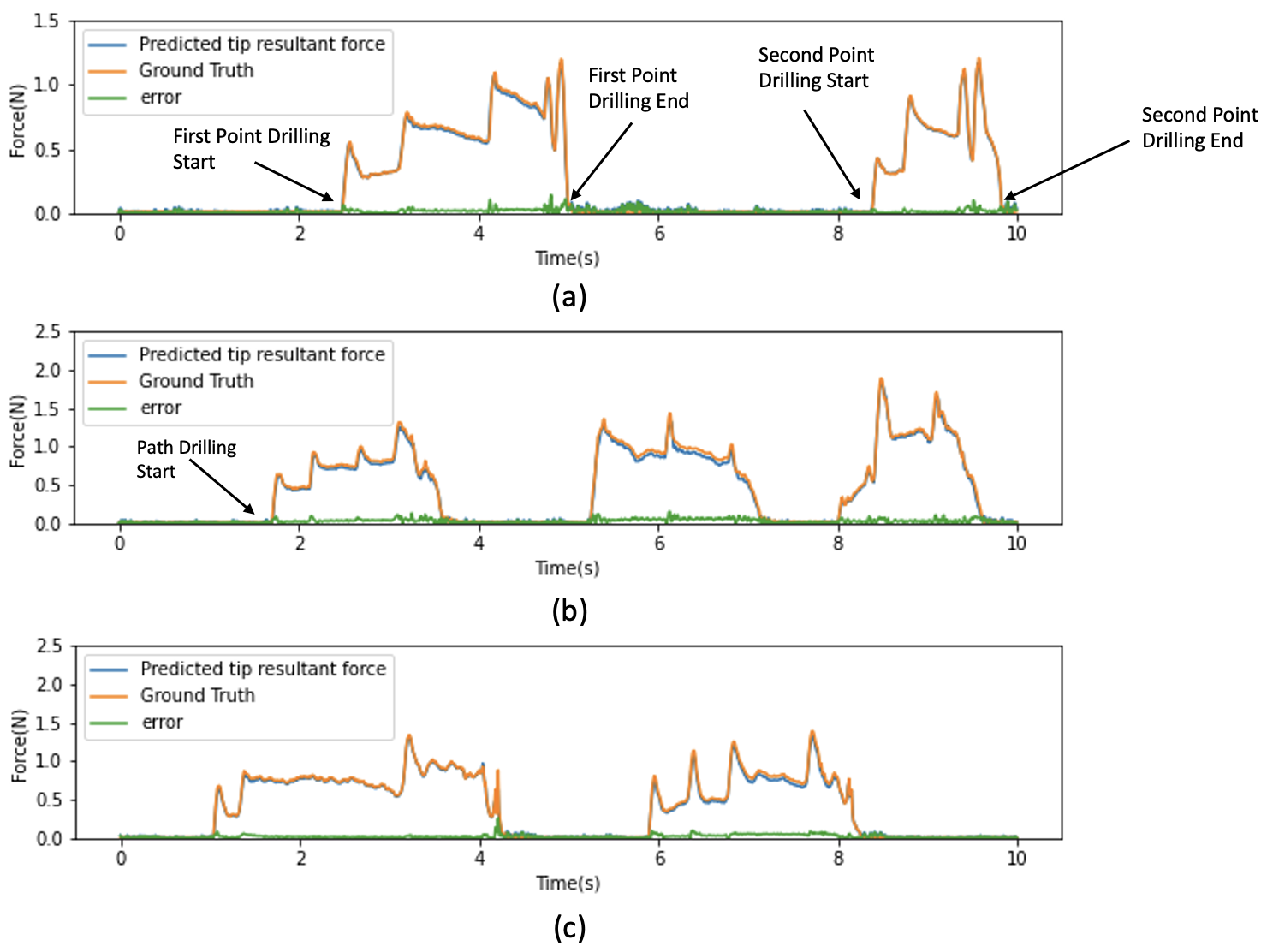}
\caption{Drilling Force Trajectory; (a) Example from Point Drilling; (b) Example Segment from Path Drilling (c) Example from Temporal Bone Phantom Drilling}\label{Drilling_force_trajectory}
\end{figure}

\begin{figure}[h]%
\centering
\includegraphics[width=0.80\textwidth]{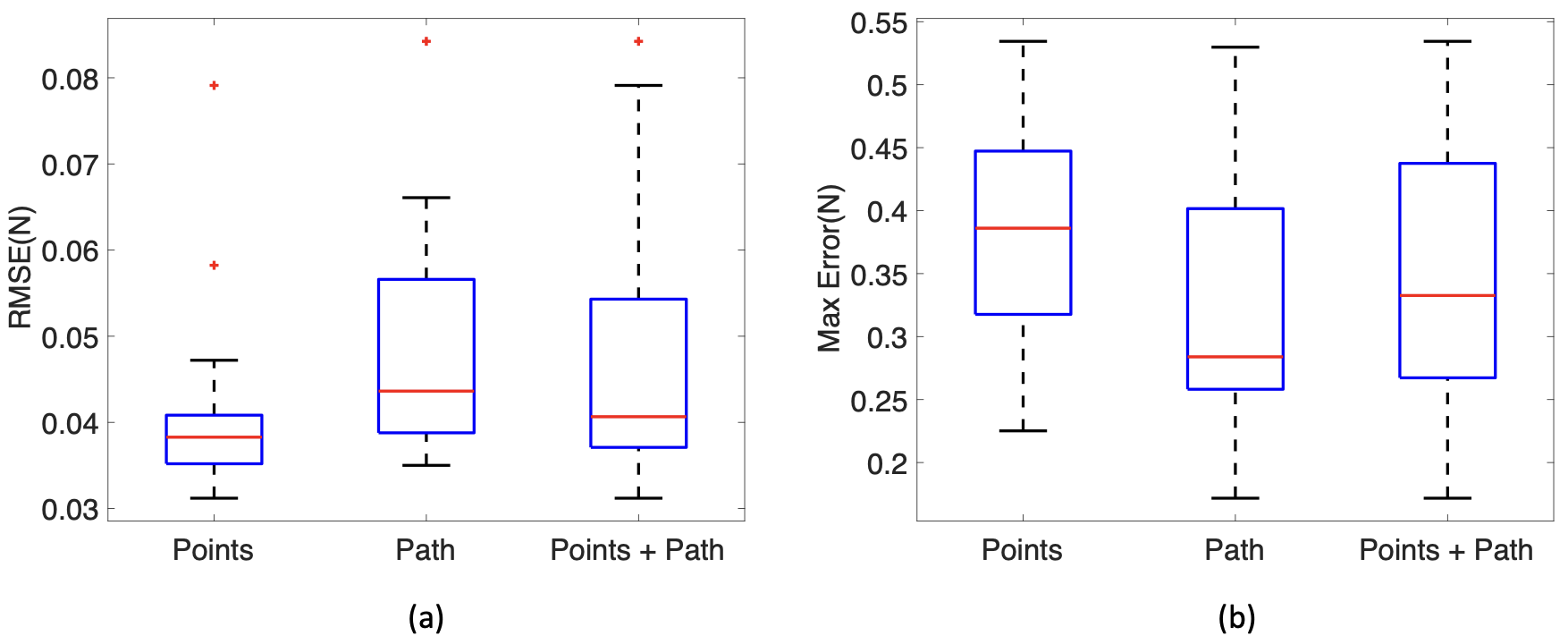}
\caption{Egg Drilling Experiments Result (N = 16 trials for both point and path drilling); (a) RMSE of N Trials; (b) Max Error of N Trials }\label{RMSE_of_Egg_Drilling_experiments}
\end{figure}

Drilling was performed on an uncooked egg to validate the use of the drill. Eggshell models are commonly used in simulated surgical training exercises for ENT due to their high fidelity to thin bone, accessibility and affordability. 
The interface of the eggshell and the outer membrane of the egg is similar to the thin, delicate bone-dural interface encountered in surgery of the temporal bone. Penetration of the egg membrane can be used as a direct metric to evaluate drilling performance. For 16 trials, the surgeon drilled the eggshell with the designed force-sensing surgical drill at nine points and one circular path as shown in Fig.~\ref{Egg_drilling_experiments_setup}b. 
Then the predicted resultant force at the drill tip was compared with the ground truth. Examples of force trajectories from point drilling and in path drilling are shown in Fig.~\ref{Drilling_force_trajectory}a and b. The RMSE between the predicted and measured values and standard deviation are shown in Fig.~\ref{RMSE_of_Egg_Drilling_experiments}. 
The minor errors during the egg drilling experiments validate the design and implementation of the instrument and calibration pipeline in the hands of a surgeon on a high-fidelity clinical model. It demonstrates the capability of real-time tip force measurement with high accuracy during the drilling of delicate structures.

\subsubsection{Comparison of Freehand and Robot-assisted Egg Drilling Force} 

\begin{figure}[h]%
\centering
\includegraphics[width=1.0\textwidth]{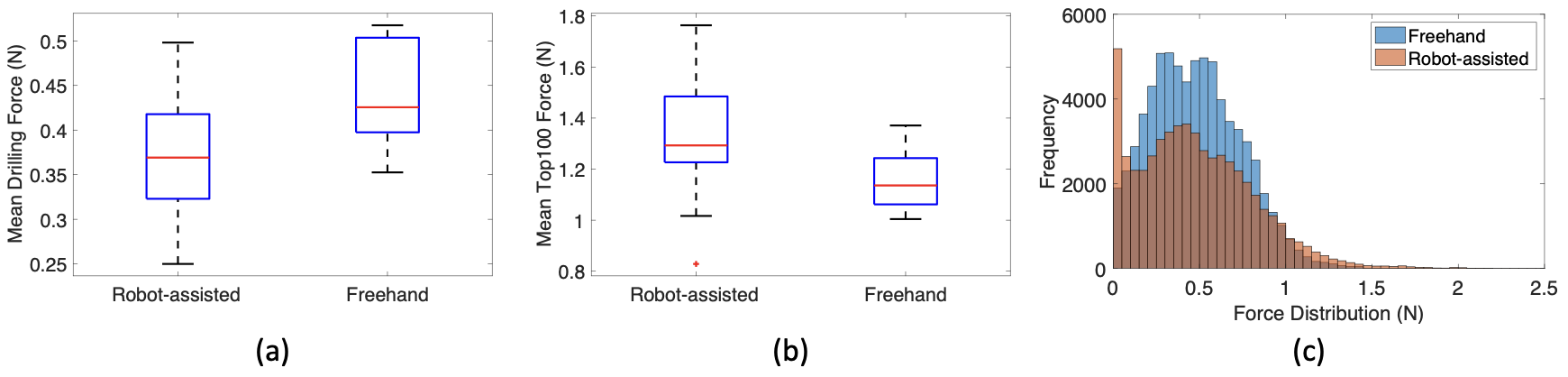}
\caption{Comparison of Robot-assisted and Freehand Drilling Force (N=10 Trials); (a) Mean Force During Drilling; (b) Mean Top 100 Force Values; (c) Distribution of Drilling Force}\label{GalenvsFreehandComparison}
\end{figure}

The forces generated by the robotically-assisted force sensing drill were compared to freehand drilling forces. For freehand drilling, the same surgeon performed the egg drilling experiment with a bare Anspach EG1 drill unattached to the robot. F/T data were collected with the ground truth phantom holder sensor for 10 trials. 
As shown in Fig.~\ref{GalenvsFreehandComparison}, the mean force and mean top 100 force values had similar ranges, indicating that the designed system generates forces similar to unassisted, freehand surgery. 

\subsubsection{Temporal Bone Phantom Drilling Experiment} 
In addition to the eggshell drilling experiments, 3D-printed temporal bone phantoms were also used to further prove the capability of the designed instrument in a wider range of applications. This task involved high-speed, coarse drilling to remove large amounts of bone. Examples of the force trajectory from temporal bone phantom drilling are shown in Fig.~\ref{Drilling_force_trajectory}c. Of 5 trials, the average RMSE was 76.6mN ($\pm$ 11.4mN) and the average maximum 100 values in all trials was 2.15N ($\pm$ 0.44N). The initial results demonstrate the potential for accurate tip forces in coarse drilling tasks with minimal error.

\section{Conclusion}\label{Conclusion}

This work reports the design and calibration for a novel force sensing otologic drill attachment for a surgical robot. The developed attachment can measure tool-to-tissue forces with millinewton accuracy in real-time with an average root mean square error of 45.8mN from 16 trials. 
Some of the measured error can be attributed to minor jitter of the robot when operated at its joint limits and the inability of the calibration model to account for inertial forces, as well as the impact of cable sag on the tool weight calibration. Ongoing work addresses these issues.
In the hands of an experienced surgeon, the forces generated by the force sensing drill are similar to freehand drilling; thus, the cooperatively controlled system and instrument design do not greatly alter the drilling task and builds confidence in the device. However, the current pipeline assumes quasi-static motion and residual error can be further reduced by implementing dynamic modeling to account for the non-negligible inertial effects.\cite{kubus2008improving,su2020deep}

Further studies include testing the accuracy of the force sensing drill on human temporal bones and heads, which will allow for evaluation of the setup on hard structures, specifically like drilling cortical bone during the initial phases of a mastoidectomy. Future work also includes integrating auditory force feedback, and implementing force-limited drilling with the robot for multiple levels of experience. Ultimately, this system has the potential to reduce trauma and complications caused by a high-speed surgical drill in a delicate, narrow corridor of the skull. Currently, an ENT surgeon requires 7-9 years of postgraduate training to operate on the skull base. This system may allow for fewer complications, improved surgical time, and increased access to care by a larger number of surgeons requiring fewer years of training to operate safely in this space. 

\backmatter

\section*{Declarations}

\bmhead{Funding} This work was supported in part by a grant from Galen Robotics, in part by Johns Hopkins University internal funds, and in part by NIDCD K08 Grant DC019708.

\bmhead{Conflict of interest} Under a license agreement between Galen Robotics, Inc and the Johns Hopkins University, Russell H. Taylor and Johns Hopkins University are entitled to royalty distributions on technology that may possibly be related to that discussed in this publication.  Dr. Taylor also is a paid consultant to and owns equity in Galen Robotics, Inc. This arrangement has been reviewed and approved by the Johns Hopkins University in accordance with its conflict-of-interest policies.


\omittext{
\textbf{Funding}: This study was funded by ...
\smallskip \\ 

\textbf{Informed consent}: This study contains data from ...
\smallskip \\ 
\textbf{Ethical approval}: This article does not contain any studies with human participants or animals performed by any of the authors.
}

\bibliography{sn-bibliography}


\end{document}